# Deep Learning Approach for Aggressive Driving Behaviour Detection

**Farid Talebloo, Emad A. Mohammed, Behrouz Far**

[farid.talebloo, eamohamm, far] @ ucalgary.ca

Department of Electrical and Software Engineering

University of Calgary, 2500 University Drive NW, Calgary Alberta T2N 1N4, CANADA[1]

[1] Acknowledgements: This research is partially supported by NSERC Discovery Grant and Alberta Major Innovation Fund (MIF).

## Abstract

Driving behaviour is one of the primary causes of road crashes and accidents, and these can be decreased by identifying and minimizing aggressive driving behaviour. This study identifies the timesteps when a driver in different circumstances (rush, mental conflicts, reprisal) begins to drive aggressively. An observer (real or virtual) is needed to examine driving behaviour to discover aggressive driving occasions; we overcome this problem by using a smartphone's GPS sensor to detect locations and classify drivers' driving behaviour every three minutes. To detect timeseries patterns in our dataset, we employ RNN (GRU, LSTM) algorithms to identify patterns during the driving course. The algorithm is independent of road, vehicle, position, or driver characteristics. We conclude that three minutes (or more) of driving (120 seconds of GPS data) is sufficient to identify driver behaviour. The results show high accuracy and a high F1 score.

## Abbreviations

DBA = Driving Behaviour Analysis

DBC = Driving Behaviour Classification

GPS = Global Positioning System

ML = Machine Learning

DL = Deep Learning

RNN = Recurrent Neural Network

GRU = Gated Recurrent Unit

LSTM = Long Short-Term Memory

## Introduction

With the number of automobile accidents, fuel economy, and determining the level of driving talent, the DBA (Driving Behaviour Analysis) becomes a critical subject to be calculated. Depending on the types of car sensors, the inputs and outputs can then be examined to establish if the DBC (Driving Behaviour Classification) is normal or deviant. According to World Health Organization (WHO) publications, studying driving behaviour is necessary. Because it is one of the primary factors contributing to catastrophe; driver factors include altitude, intoxication, fatigue, poor road conditions, eyesight impairment, and vehicle performance considerations [1]. Thus, considering the amount of damage in a car accident, in this study, we examine the problem of detecting violent driving using GPS data with a few minutes of driving per person.

To monitor driving behaviour, many types of sensors connected to a control area network may be employed. Driving behaviour data is multidimensional, time-series data that has been calculated. In some cases, the dimensions of time series data are not statistically independent [2]. A proper depiction of driving characteristics might be critical in applications such as autonomous driving, auto insurance, and others. On the other hand, traditional methods rely significantly on handcrafted features, impeding ML algorithms' potential to reach superior efficiency [3]. Driving is a dynamic endeavour requiring various ability levels (e.g., acceleration, braking, turning); this dynamicity can distinguish drivers' driving behaviour. Compared to a fingerprint, everyone has distinct driving patterns, such as a set speed, acceleration, and braking patterns [4]. However, there are different ways to find patterns and classify them in particular [5], [6].

Driving behaviour analysis will assist us in determining driver efficiency, enhancing traffic safety, and eventually encouraging the development of intelligent and resilient transportation systems. Although various attempts have been made to evaluate driving behaviour, representation learning can improve current methodologies by exploring peer and temporal connections in driving behaviour [7].

Numerous reports on road safety have focused mainly on the elements that contribute to severe and fatal accidents. As a result, less emphasis has been placed on minor injuries or events before a crash [8]. This reality may lead to inaccurate beliefs about injury prevention and management. While the DBC received considerable attention in the past, much remains to be learned about it, including the dimensions of driving patterns and their potential impact on road safety [9].

## Scope of research

This study proposes a model that can discover the aggressive driving pattern in less than 3 minutes (the selection of 180 seconds will be described later in this research) by capturing GPS records every second. We examine two different RNN-based methods (GRU, LSTM) in various circumstances. The experiments have been evaluated in two approaches: 1) Splitting the dataset into training, validation, and test datasets; and 2) reserving a driver dataset to have the real-world test.

## Driving Behaviour

Because of different driving factors (fatigue, intoxication, drowsiness, distraction), drivers' behaviour may vary; road adhesion, traffic, and weather conditions also influence driving traits [10]. The choice of driving speed is one element of driving style that has emerged as a significant predictor of differential accident participation in recent years [11]. Unsafe driving habits may develop because of two main factors. First and foremost, drivers may have varying attitudes about driving, including varying anxiety levels about the potential of a collision. Second, drivers may have differing perspectives on what constitutes good and poor driving and their level of competence and safety on the road [5]. Although there is a link between some demographic characteristics and accident risk, this link is mediated by several other variables that should be considered. Despite this, age and gender are related to accident risk even after considering these driving style variables [12]. In this study, we want to find the correlation between changes in vehicle states and driving behaviours. In the following section, we will discuss different sensors that may be utilized to find the aforementioned relation.

## Sensors

An observer (online = human, fleet tracker / offline = sensors, camera records) must evaluate their driving conduct in a driving course. There are proposed models that try to monitor the drivers by utilizing various sensors and detecting driving behaviour by changing drivers' different conditions [13] [14]. There are two types of sensors in autonomous and non-automated vehicles: "Environmental" and "Vehicle State" [15]. In this study, we are using the vehicle states (position and changes) sensors. The most applicable vehicle states sensors are: IMU, CAN J1939, Magnetic Compass and GPS. Regarding the dataset used in this study, we will study GPS deeper in the section that follows.

## GPS

The GPS does not require data transmission from the user; it functions independently of any cellphone or internet reception [16]. GPS is crucial for military, civil, and commercial purposes worldwide [16]. The US government developed, maintains, and makes the system openly available to anybody with a GPS device [16].

The GPS receiver determines its position and time using data from many GPS satellites. This data is sent to the receiver by each satellite. A very stable atomic clock, synced with ground clocks, is carried by each satellite. Day-to-day corrections are made for time differences. Similarly, the satellite placements are precise[17]. GPS receivers also contain clocks, but they are less accurate and steady. Radiation delays are proportionate to distance since radio waves have a fixed speed regardless of satellite speed. Because the receiver must compute four unknown values, four satellites are required at minimum (three position coordinates and a clock deviation from satellite time)[17].

The raw GPS record captured from the mobile devices has at least these attributes: Speed (Km/h), Latitude, Longitude, Altitude, Vertical accuracy, and Horizontal accuracy. This article attempts to classify driving behaviour using most of these parameters first and subsequently only using speed variations and their correlation with latitude and longitude variations.

## Classification approaches

This step uses GPS (Latitude, Longitude, Altitude) data to classify aggressive or non-aggressive driving behaviour. As the trajectory data of a driver is a sequential dataset (time series), we use RNN approaches in this study [18]. RNN is a subtype of supervised DL in which the previous step's output is used as input for the next phase. For sequential data, the RNN deep learning method is optimal [18]. The hidden state, which memorizes certain information about a sequence, is the most crucial aspect of RNN. RNN transforms independent activations into dependent activations, decreasing the complexity of increasing parameters and remembering each previous output by sending each output to the next hidden layer as input. In this study, we go through LSTM and GRU methods to experiment with the accuracy of the proposed model in different circumstances. In the following section, the differences between GRU and LSTM are described.

## GRU vs. LSTM

RNNs are designed to work with time series; they use previous sequence information to produce current output. Memory problems arise in RNNs because of a vanishing gradient. RNN suffers from vanishing gradients more than other neural networks as the number of steps increases[19]. To explain vanishing gradients, consider the following procedure: to train an RNN, we backpropagate through time, computing the gradient at each step. The gradient is used to update the weights of the network. If the previous layer's effect on the current layer is modest, the gradient value will be low. If the gradient of the previous layer is slight, the gradient of the following layer will be as well. Gradients grow smaller as we backpropagate. A smaller gradient indicates that there will be no weight update. Consequently, the network fails to learn prior inputs, causing short-term memory problems [20].

To solve the vanishing gradients issue, two customized variants of RNN were developed. They are as follows: 1) GRU and 2) LSTM. LSTMs and GRUs use memory cells to store the activation value of preceding steps in a long sequence. Gates are used in networks to regulate the flow of information. Gates can learn which inputs in a sequence are significant and store their knowledge in a memory unit. They may provide data in lengthy sequences and utilize it to generate predictions [21].

The workflow of GRU is the same as RNN, but the difference is in the operations inside the GRU unit. Inside GRU is two gates: 1) reset gate and 2) update gate. Gates are nothing but neural networks; each gate has its weights and biases. The update gate decides if the cell state should be updated with the candidate state (current activation value) or not. The reset gate is used to decide whether the previous cell state is essential or not. The candidate cell is simply the same as the hidden state (activation) of RNN. The final cell state is dependent on the update gate. It may or may not be updated with the candidate state. The final cell Removes some content from the last cell state and writes some new cell content [20].

Long short-term memory (LSTM) is like GRU in that they are designed to address the vanishing gradient issue. In addition to GRU, there are two additional gates here: 1) the forget gate 2) the output gate. Because all three gates use the sigmoid activation function, they are between 0 and 1. The forget gate determines what is retained and forgotten from the previous cell state; it determines how much information from the previous state should be kept and how much should be lost. The output gate determines which portions of the cell are sent to the concealed state [20].

## Dataset and challenges

The "UAH-DriveSet" [22] is a public repository of data collected from various testers in various conditions by the driving monitoring program "DriveSafe." This dataset intends to speed progress in driving analysis by providing many characteristics collected and analyzed during independent driving tests using all smartphones' sensors and capabilities. The application was run on six distinct drivers and vehicles. At the same time, they engaged in three distinct behaviours (normal, drowsy, and aggressive) on two distinct types of roads (highway and secondary road), yielding over 500 minutes of naturalistic driving with associated raw data and additional semantic information, as well as video recordings of the trips [22]. This dataset contains six distinct drivers; each driver drove two roads - one of which was 25 kilometres long and with a maximum speed limit of 120 kilometres per hour, and the other of which was 16 kilometres long and had a maximum speed limit of 90 kilometres per hour - while exhibiting three distinct driving behaviours: normal, drowsy, and aggressive. Because the "RAW GPS.txt" file contains nearly all the information we want, including speed, latitude, longitude, altitude, vertical and horizontal precision, and sample timestamp. We use it to categorize the drivers' behaviour. While this is the ideal dataset for our situation, it does have some limitations. The drivers are comprised of five men and one female. The minimum and maximum age ranges are 20-30 and 40-50. Except for one car, they are all made in Europe. Additionally, one of the cars is a battery-electric vehicle. In the next section, we examine the limitations and the solutions that we propose to mitigate them.

| Driver | Gender | Age | Vehicle | Fuel type |
|--------|--------|-----|---------|-----------|
| D1 | Male | 40-50 | Audi Q5 (2014) | Diesel |
| D2 | Male | 20-30 | Mercedes B180 (2013) | Diesel |
| D3 | Male | 20-30 | Citroen C4 (2015) | Diesel |
| D4 | Female | 30-40 | Kia Picanto (2004) | Gasoline |
| D5 | Male | 30-40 | Opel Astra (2007) | Gasoline |
| D6 | Male | 40-50 | Citroen C-Zero (2011) | Electric |

*Table 1 List of drivers and vehicles that performed the tests*

## Dataset limitations and solutions

1. Each driver's behaviour differs from the others; for example, one driver's normal behaviour may differ from others. As a solution, we have all the drivers' training data by splitting a trip into smaller batches and giving some to the system for learning and the rest for validation and testing.
2. The most challenging aspect of the dataset is labelling an entire route with a single style of driving. However, there have been instances when the driver drove contrary to the label. We choose to ignore it as noise; given the nature of human beings, it is quite natural for an individual's behaviour to vary throughout a driving course.
3. Because each driver's driving speed varies, the quantity of data points in each dataset file varies. The solution is to divide our dataset into smaller groups, work on them individually, and make the frequency of the events the same.
4. To feed our data to the model, the batches must have the exact dimensions. For example, we cannot provide 20 kilometres of one route for learning in one batch and the remaining 5 kilometres (3 kilometres for testing and 2 kilometres for validation) in two separate batches. The solution is to divide our dataset into smaller batches, the same size as the data points, and work on them individually.
5. The low volume dataset for deep learning is an issue, as the dataset content described previously. Obviously, for deep learning pattern recognition, we need to have more volume of the dataset. So, we proposed overlapped time series sequences to provide more training, validation, and test dataset as input for the designed model.
6. Model sensitivity occurs during the model training because of imbalanced weights of classes. The nature of aggressive or non-aggressive driving made this an issue. It means that when we drive, the potential of aggressive driving is far less than normal driving. We oversampled aggressive driving during the training and validation phase and used weighted classes in our RNN models to address this phenomenon.

## Methodology

Our proposed model is a labelled (supervised) data-based deep learning pattern recognition model. The best outcome is then reached in the evaluation (real-world test) process by adjusting the topology and hyperparameters. The time-series data of speed changes in a driving trajectory is shown in the figure-1. In addition to speed, our dataset also includes geographic position information, which leads to a better outcome in the deep learning process.

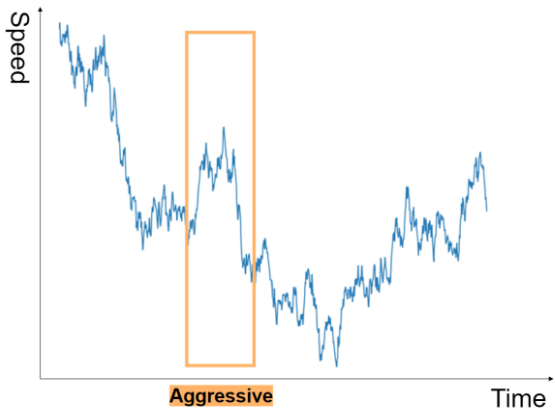

*Figure 1 Speed changes of a trajectory*

We construct various LSTM and GRU architectures with varying parameter values and evaluate their output to determine which design best fits our situation. The original dataset is divided into three sections: training, validation, and test. Two types of normalization (Min-Max and Standardization) approach are used first, followed by a seven-layer LSTM/GRU model with dropout, batch normalization, delta, and shuffled training datasets before splitting.

Figure 2 shows that in the first layer, we are preparing a trajectory value of a driver with 120 seconds of their driving that each moment has eight features (speed, longitude, latitude, altitude, and differentials). Then, the following seven neural networks are LSTM/GRU (different experiments), drop out (make it more complex for the model to find patterns easier), and a batch normalizer that helps to the consistency of values sent to the next layer. After seven layers of LSTM/GRU layers, a dense layer with activation of sigmoid classify the pattern of the trajectory to one of the classes of aggressive or non-aggressive.

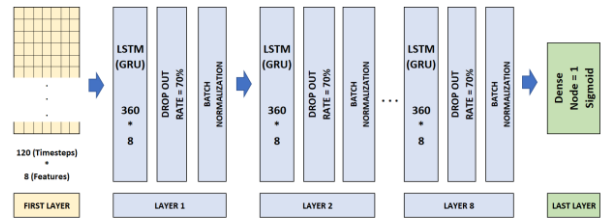

*Figure 2 Neural Network Diagram*

We implement 24 different models based on the parameters below:

1. **Input shape:** The input shape went from 60 seconds to 120 and then to 180, which improved the accuracy; the trend shows that we will get better accuracy than driving GPS records to be sent for the input layer. Nevertheless, as we want to implement a real-world model to be deployable on an embedded device, the accuracy (98% on test evaluation) of the model that works with 2 minutes of driving would be sufficient.

2. **Normalization type:** Normalization is used to reduce the amount of duplication in a connection or collection of relationships. In the experiments, the "Standardization" was shown to be superior to "Min-Max" normalization.

3. **The number of Hidden Layers:** To obtain the optimal decision boundary, we must use hidden layers; thus, we gradually increase the number of hidden layers from one to seven, with seven providing the best results. We make the model complex enough to capture all the patterns and map them to the correct target class.

4. **Dropout:** Dropout is a technique for preventing a model from overfitting, and so introducing dropout after each GRU/LSTM layer helped our model avoid overfitting. We use a 70% dropout value to let the model find all its nodes, relations, and weights.

5. **Batch normalization:** Batch normalization is a technique for standardizing the inputs to a network that can be applied to either the activations of a preceding layer of inputs or to the activations of a subsequent layer of inputs directly. Batch normalization accelerates training in some situations by halving or bettering the number of epochs and provides some regularization, hence lowering generalization error. Batch normalization undoubtedly aids our model.

6. **Shuffling the training Dataset:** After several tests, we discovered that one of the issues with model training is that the last segment of each trajectory is always given to the model for validation. However, the model picked up on 70% of drivers' first behaviour. Nevertheless, the model will encounter all drivers' behaviour using this new method, which involves shuffling the training dataset before dividing it.

7. **Adding differentials:** The delta rule is a gradient descent learning rule for updating the weights of the inputs to a neural network. It is a particular case of the more general backpropagation algorithm and helps in slightly improving the accuracy.

## Data preprocessing

1. Single row as a data point vs time-series: If we treat each row as a data point in the primary technique and pass them to our model, we notice that this strategy is invalid for our problem because the data points are related in time. It is illogical to classify a driver's behaviour based on a single geographical data point. For example, if only an individual driver is in a specific location, we cannot judge that they are driving aggressively.

2. Window size of data points, time-correlation: We know our dataset is a time series dataset since each row corresponds with the preceding and next rows. Additionally, we understand that the driver's current location is irrelevant at any given time, but the location changes and how they occur are critical to us. As a result, we chose to employ the window size notion, treat a batch of rows as a single data point, and send them to the model to verify that it is discovering relationships between distinct data points.

3. Variation and road-type features: We enhance our model by including variations and road-type variables. Variations may be a more helpful feature for our model because they connect two distinct data points and aid the model in determining their correlation. Additionally, given that we have two distinct roads, understanding the road type can aid the model. Nevertheless, we try to add those variations to make the dependency of the model to the specific location of the road that the dataset gathering occurs, and it drastically helps the proposed model.

4. Shuffling: We choose various segments of each road as a test and validation dataset to demonstrate the route's whole course to our model. For instance, rather than utilizing the first 20 kilometres of our road for training and the remaining five kilometres for validation and testing (on a 25-kilometre road), we split our dataset into the specified window size first, then mixed the data and utilized a percentage of batches for testing.

5. Overlapping the datapoints: When we utilized a larger window size, for example, 180 sequences of GPS records, the number of data points decreased substantially. Thus, we applied the overlapping idea. In this study, we use the sliding window concept to choose the batches.

6. The "speed limit" as a feature: We included a speed difference with the road's maximum speed to the dataset because we have two roads with varying maximum speeds in our dataset. It enables the model to link the top speed with driving behaviour.

7. Evaluation based on "not seen driver" (real-world evaluation): Since our dataset has six distinct drivers, we partition it to make our model more generic. Five drivers are chosen for training and validation, and one is separated and never shown to the model to reserve for the testing phase. This practice simulates real-world tests conducted in our model evaluation section.

By more understanding of the data, it is possible to enhance the subsequent stages of DL activities. Comprehending data implies the completeness of the data, its purpose and application[19]. Then comes the data cleaning phase, which entails completing gaps in data, smoothing out noise, identifying and removing outliers, and addressing discrepancies. Following that, data integration is required, which is often necessary when combining different databases or files. Each driver's dataset occupies a separate folder, and each folder contains different types of driving on different roads. So, in this step, we code a function to walk between folders and combine all the files into one integrated file with the tags of driver label, type of driving and name of the trajectory.

In the following data conversion, we are confronted with data normalization, modification, and aggregation procedures at this data preprocessing step. We use both Min-Max scaling and standardization in preprocessing of feature scaling of data for the model. A critical finding of Min-Max Scaling is that it is strongly affected by our data's highest and lowest values, which means it will be skewed if our data includes outliers. So, all experiments that utilize standardization have better results. In the following, there are two examples and charts of the feature scaling concept.

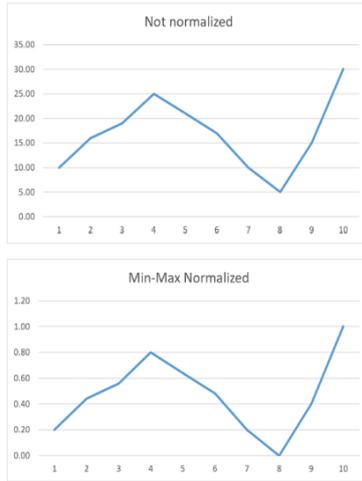

Figure 3 Min-Max Normalization formula and example

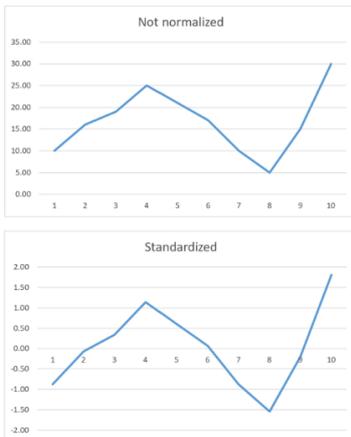

Figure 4 Standardization formula and example

## Split dataset preparation

Dataset Split is a technique to evaluate the performance of the ML classification model. We take a given dataset and divide it into three subsets. The training dataset is a set of data used for learning (by the model), that is, to fit the parameters of the ML model [23]. The validation dataset is a set of data used to provide an unbiased evaluation of a model fitted on the training dataset while tuning model hyperparameters. It also plays a role in other forms of model preparation, such as feature selection and threshold cut-off selection. The testing dataset is a set of data used to provide an unbiased evaluation of a final model fitted on the training dataset [19].

Our proposed model for the learning phase experiences two different methods. In the first method, shown in Figure 2, all the information is first merged. Then the training, validation and test dataset are split. In this experiment, the model has a chance to see at least part of the trajectory of all drivers. However, in the second method, the D5 driver is wholly excluded from the training and validation process and is only evaluated during the model testing. The advantage of the first method is that the model faces more datasets; hence the disadvantage is that the model may have poorer performance in the real world. The benefit of the second method is that the simulation model has practically passed the real-world test. However, at the same time, it loses the chance of encountering a more significant data set during training, which will be solved during deployment. The way to solve the last problem is that the model is simultaneously in classifier and learning mode in the production environment, which also classifies driving behaviour. Moreover, it learns new scenes.

| Dataset name | Percentage (of selected datasets) | Data of Drivers (shuffled) | | | |
|---|---|---|---|---|---|
| Training | 80% | D1 | D2 | D3 | D4 | D6 |
| Validation | 20% | | | | |
| Testing | 100% | D5 | | | |

Table 2 Unseen driver for the test phase

| Dataset name | Percentage (of selected datasets) | Data of Drivers (shuffled) | | | | | |
|---|---|---|---|---|---|---|---|
| Training | 70% | D1 | D2 | D3 | D4 | D5 | D6 |
| Validation | 15% | | | | | | |
| Testing | 15% | | | | | | |

Table 3 Normal approach of dataset split

## Changes of the values

One of our primary responsibilities as model designers is to minimize the model's reliance on the dataset. As a result, we added modifications to each dataset record's GPS values, such as latitude, longitude, and altitude, in this phase. Thus, the first record of each route is treated as 0. Furthermore, each timestep tracks changes. For example, in Figure 3, it is shown that the latitude value changes from 23.45 to 29.45. Thus, in the "latitude delta" cell, positive six changes of values are logged.

| Timestep | Latitude | Longitude | Altitude |
|---|---|---|---|
| 1 | 23.45 | 56.78 | 12.56 |
| 2 | 29.45 | 59.78 | 11.50 |
| … | … | … | … |

↓

| Timestep | Latitude | Longitude | Altitude | Latitude Delta | Longitude Delta | Altitude Delta |
|---|---|---|---|---|---|---|
| 1 | 23.45 | 56.78 | 12.56 | 0 | 0 | 0 |
| 2 | 29.45 | 59.78 | 11.50 | +6 | +3 | -1.06 |
| … | … | … | … | … | … | … |

Figure 5 Changes of values

## The overlapped divided method

The UAH database contains various trajectories of six drivers on two distinct roads while exhibiting various driving behaviours. If we utilize basic cuts to identify driving patterns, our data volume will significantly decrease. If we create trajectories using the overlapping split method, our data volume will grow by up to tenfold. In another way, since our suggested model is based on deep machine learning techniques, it needs a more significant amount of data. In Figure 4, it is evident that with 6 points of timesteps instead of two trajectories, we use them in 6 trajectories.

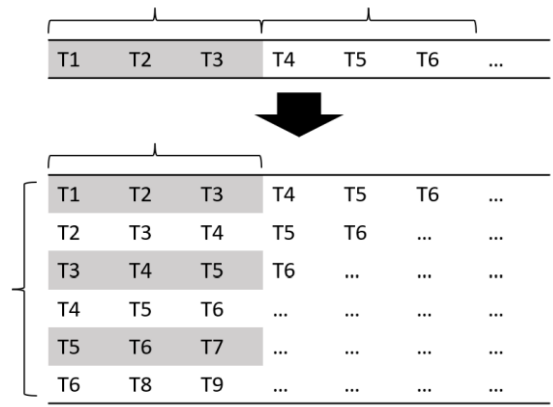

Figure 6 Overlapping of trajectories timeseries

## Evaluation

Once we fit a deep learning neural network model, we must assess its performance on a test dataset [23]. Evaluation of the Test dataset is crucial since the reported performance lets us pick between candidate models and educates us about how optimum the model is at handling the problem. We employ a conventional binary classification issue in this work.

## Accuracy

The difficulty with accuracy as our primary performance metric is that it does not fare well with a significant class imbalance. Let us explain our model parameters in our target vector, 0 signifies normal, and 1 means aggressive driving. If we utilize the accuracy statistic, it says that 99 percent of the time, the model characterizes the "normal" driving accurately. However, when it comes to identifying aggressive driving, it can categorize 70 percent of inputs. So, the accuracy cannot reflect an acceptable metric for our model. In the following section, we will explain the "F1 Score".

## F1 Score

The F1 measurement is an overall assessment of a model's correctness that combines precision and recalls in that addition and multiplication blend two compositions to form a different notion entirely [24]. An excellent F1 score indicates we have low false positives and low false negatives, so we properly detect genuine threats and are not bothered by false alarms. An F1 score is ideal for 1, whereas the model is a dismal failure with 0.

All models will create some false negatives, some false positives, and maybe both. While we may tweak a model to reduce one or the other, we typically confront a trade-off, where a decrease in false negatives increases false positives or vice versa. We will need to optimize for the performance measures that are most beneficial for our unique situation.

## Experimental results

The findings (Figure 5) indicated that with the LSTM algorithm, we had an accuracy of 99.6 percent in three minutes of driving and 98.4 percent in two minutes of driving. Due to the near distance between three minutes and two minutes, we might offer the model based on two minutes between two to three minutes.

The characteristics and settings of the model are detailed below. The number of input ports is 120 nodes, and the number of intermediate levels is eight, with each layer containing 360 nodes. In the last layer, we employed a layer with a Sigmoid activation function to conduct the categorizing. In the last layer, we utilized the output bias initializer to alter the weight between classes.

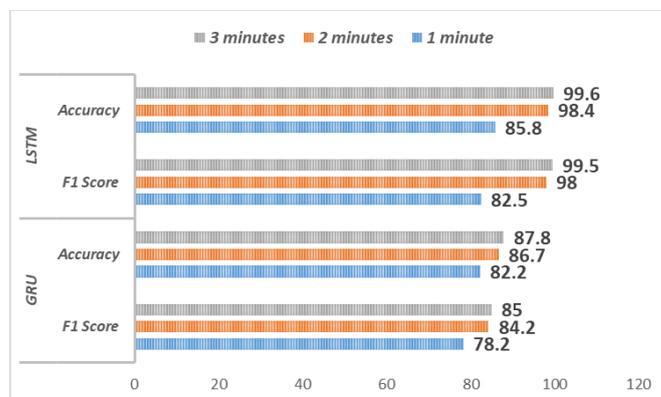

Figure 7 Models Evaluations

In Figure 6, we can see the accuracy and F1 score values. These tests are generated by entirely reserving the dataset of a driver (Driver 5) for the validation phase. Our suggested model achieves 93 percent accuracy in real-world testing with three minutes of driving. In prior investigations, this approach has not been employed for validation and is essential in its type.

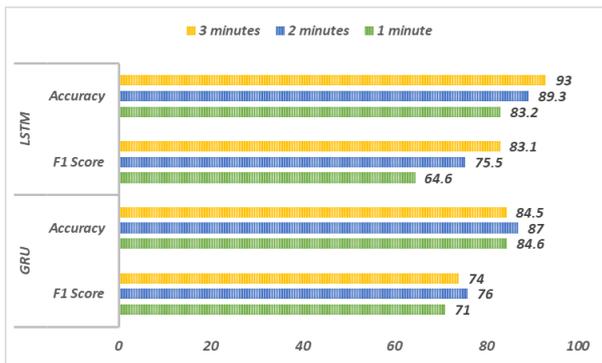

Figure 8 Real-world Evaluations

## Conclusion

Our dataset comprises time-series data, and algorithms that can detect the relationship between time and feature changes produce excellent results in our study, as shown in table 6. Regardless of whether the model is evaluated using reserved validation data or real-world tests. The real-world results also show that their driving behaviour can be characterized (aggressive or non-aggressive) by having GPS recordings of around three minutes of the driving trajectory.

## Future works

This model will be embedded in a program run on multiple smartphone operating systems in the following phases. Because the suggested AI model can identify the driver's behavioural driving behaviour every three minutes, the information is synced every three minutes with the server system situated in the cloud. This device will function with any cellphone that permits access to GPS, regardless of the type of vehicle and whether the vehicle is a self-driving or conventional type.

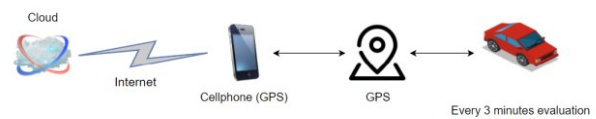

Figure 9 Embedded AI Model in the smartphone

## Appendix:

| NN | timestep | normalization | Evaluation | loss | accuracy | precision | recall | F1 Score |
|---|---|---|---|---|---|---|---|---|
| GRU | 60 | Min-Max | Seen | 0.5617 | 0.776 | 0.687 | 0.822 | 0.748461233 |
| GRU | 60 | Standardization | Seen | 0.4212 | 0.822 | 0.775 | 0.789 | 0.78193734 |
| GRU | 120 | Min-Max | Seen | 1.0759 | 0.781 | 0.736 | 0.714 | 0.724833103 |
| GRU | 120 | Standardization | Seen | 0.379 | 0.867 | 0.81 | 0.877 | 0.842169532 |
| GRU | 180 | Min-Max | Seen | 0.8107 | 0.736 | 0.689 | 0.629 | 0.657634294 |
| GRU | 180 | Standardization | Seen | 0.4329 | 0.878 | 0.843 | 0.857 | 0.849942353 |
| GRU | 60 | Min-Max | Unseen | 0.2487 | 0.896 | 0.805 | 0.711 | 0.755085752 |
| GRU | 60 | Standardization | Unseen | 0.3219 | 0.846 | 0.619 | 0.83 | 0.709137336 |
| GRU | 120 | Min-Max | Unseen | 0.5059 | 0.801 | 0.533 | 0.802 | 0.640398502 |
| GRU | 120 | Standardization | Unseen | 0.3903 | 0.87 | 0.643 | 0.927 | 0.759313376 |
| GRU | 180 | Min-Max | Unseen | 0.4403 | 0.828 | 0.58 | 0.789 | 0.668546384 |
| GRU | 180 | Standardization | Unseen | 0.2088 | 0.845 | 0.588 | 0.991 | 0.738072198 |
| LSTM | 60 | Min-Max | Seen | 0.3348 | 0.822 | 0.762 | 0.816 | 0.788076046 |
| LSTM | 60 | Standardization | Seen | 0.2913 | 0.858 | 0.821 | 0.829 | 0.824980606 |
| LSTM | 120 | Min-Max | Seen | 0.2997 | 0.923 | 0.963 | 0.841 | 0.897874723 |
| LSTM | 120 | Standardization | Seen | 0.0544 | 0.984 | 0.973 | 0.987 | 0.97995 |
| LSTM | 180 | Min-Max | Seen | 0.0602 | 0.981 | 0.974 | 0.98 | 0.976990788 |
| LSTM | 180 | Standardization | Seen | 0.0124 | 0.996 | 0.994 | 0.997 | 0.99549774 |
| LSTM | 60 | Min-Max | Unseen | 0.4245 | 0.837 | 0.632 | 0.664 | 0.647604938 |
| LSTM | 60 | Standardization | Unseen | 0.3991 | 0.832 | 0.619 | 0.675 | 0.645788253 |
| LSTM | 120 | Min-Max | Unseen | 0.7365 | 0.86 | 0.667 | 0.727 | 0.695708752 |
| LSTM | 120 | Standardization | Unseen | 0.5766 | 0.893 | 0.764 | 0.746 | 0.754892715 |
| LSTM | 180 | Min-Max | Unseen | 0.4201 | 0.903 | 0.771 | 0.778 | 0.774484183 |
| LSTM | 180 | Standardization | Unseen | 0.4131 | 0.93 | 0.855 | 0.809 | 0.831364183 |

*Table 4 Experiments details*